%% file: arxiv.tex
\documentclass[sigconf,natbib=false]{acmart}

\AtBeginDocument{%
  }

\copyrightyear{2026}
\acmYear{2026}
\setcopyright{cc}
\setcctype{by}
\acmConference[CIKM '26]{Proceedings of the 35th ACM International Conference on Information and Knowledge Management}{November 07--11, 2026}{Rome, Italy}
\acmBooktitle{Proceedings of the 35th ACM International Conference on Information and Knowledge Management (CIKM '26), November 07--11, 2026, Rome, Italy}
\acmDOI{10.1145/3799682.3840211}
\acmISBN{979-8-4007-2539-5/2026/11}
\acmDataLink{https://zenodo.org/records/22077479}
\acmCodeLink{https://github.com/National-Running-Club-Database/data_paper}

\RequirePackage[
  datamodel=acmdatamodel,
  style=acmnumeric,
]{biblatex}
\usepackage{tikz}
\usetikzlibrary{arrows.meta, positioning}
\usepackage{placeins}

\begin{document}

\input{cikm_body}

\printbibliography

\appendix
\FloatBarrier
\input{appendix}

\end{document}

%% file: cikm_body.tex

\title{NRCD: An Open Database of Collegiate Running with Unified Performance Standardization}

\author{Jonathan A. Karr Jr}
\email{jkarr@nd.edu}
\email{nationalrunningclubdatabase@gmail.com}
\affiliation{%
  \institution{University of Notre Dame}
  \city{Notre Dame}
  \state{Indiana}
  \country{USA}
}
\affiliation{%
  \institution{National Running Club Database}
  \state{Michigan}
  \country{USA}
}

\author{Ryan M. Fryer}
\email{jpv8cu@virginia.edu}
\affiliation{%
  \institution{University of Notre Dame}
  \city{Notre Dame}
  \state{Indiana}
  \country{USA}
}
\affiliation{%
  \institution{University of Virginia}
  \city{Charlottesville}
  \state{Virginia}
  \country{USA}
}

\author{Ben Darden}
\email{bdarden1205@vt.edu}
\affiliation{%
  \institution{Virginia Tech}
  \city{Blacksburg}
  \state{Virginia}
  \country{USA}
}

\author{Nicholas Pell}
\email{npell@alumni.unc.edu}
\affiliation{%
  \institution{University of North Carolina at Chapel Hill}
  \city{Chapel Hill}
  \state{North Carolina}
  \country{USA}
}

\author{Kayla Ambrose}
\email{kambrose@alumni.nd.edu}
\affiliation{%
  \institution{University of Notre Dame}
  \city{Notre Dame}
  \state{Indiana}
  \country{USA}
}

\author{Evan Hall}
\email{ehall9@alumni.nd.edu}
\affiliation{%
  \institution{University of Notre Dame}
  \city{Notre Dame}
  \state{Indiana}
  \country{USA}
}

\author{Ramzi K. Bualuan}
\email{rbualuan@nd.edu}
\affiliation{%
  \institution{University of Notre Dame}
  \city{Notre Dame}
  \state{Indiana}
  \country{USA}
}

\author{Nitesh V. Chawla}
\email{nchawla@nd.edu}
\affiliation{%
  \institution{University of Notre Dame}
  \city{Notre Dame}
  \state{Indiana}
  \country{USA}
}

\renewcommand{\shortauthors}{Karr et al.}

\begin{abstract}
  Collegiate running in the United States generates thousands of race results annually in cross country and track and field, yet no large-scale dataset has been publicly available for research. Existing websites such as Athletic.net, MileSplit, and TFRRS host results but do not support bulk download, restricting prior analyses to \textasciitilde{}500 performances, often skewing studies toward male athletes. We introduce the \textbf{National Running Club Database (NRCD)}, the first openly available collegiate running dataset at scale: 143,868 approved performances from 31,351 athletes across 1,423 meets in four sports (cross country (XC), indoor and outdoor track, and road races), 36.2\% women, spanning 2003-2026. Meets from August 2023 onward carry comprehensive course distance, elevation gain and loss, weather at race time, and track venue metadata (99.9\% of XC rows with weather fields). NRCD is \textbf{community-governed} through open submission and expert approval and is maintained as a live database whose meet volume has grown yearly. We release a unified performance standardization framework that operationalizes established distance, elevation, and heat adjustments in one pipeline; \emph{XC-only validation}; heat is a Hadley-band heuristic. \textbf{We recommend gender-stratified modeling}. On XC, full standardization lowers median within-athlete cross-meet variability by 51.1\% (women) and 35.4\% (men) versus raw times. We release the dataset and pipeline with a Python package `nrcd' under FAIR principles, supporting longitudinal athlete modeling, environmental-confounder studies, and gender-equity research in collegiate sport.
\end{abstract}

\ccsdesc[500]{Information systems~Data mining}
\ccsdesc[300]{Applied computing~Health informatics}

\keywords{sports analytics, running, performance standardization, longitudinal modeling, community-contributed data, collegiate athletics}


\maketitle
Dataset with Datasheet: \url{https://zenodo.org/records/22077479}

\noindent Code:  \url{https://github.com/National-Running-Club-Database/data_paper}

\noindent Package: \url{https://pypi.org/project/nrcd/}

\noindent GitHub for Python package: \url{https://github.com/National-Running-Club-Database/nrcd}

\noindent Website:  \url{https://www.nationalrunningclubdatabase.com}
\section{Introduction}
Collegiate distance running is organized across World Athletics, NCAA, NAIA, NJCAA, and club bodies including the National Intercollegiate Running Club Association (NIRCA)~\cite{nirca}. Officiated competition in NIRCA ranges from NCAA Division~I-caliber performances to times of community 5K road races, so an open resource must serve a full band of runners rather than a single subfield or ability tier.  Thousands of results are published annually on Athletic.net, MileSplit, and TFRRS~\cite{athleticnet,milesplit,tfrrs}, yet none of these platforms support bulk relational export. Prior open analyses of running performance have therefore relied on hand-curated samples of a few hundred records~\cite{millettpredicting}, with documented male skew~\cite{anderson2023under,james2023underrepresentation,smith2022auditing}. A separate barrier is \emph{comparability}: cross country courses differ in measured distance, terrain, and weather; track meets differ in surface, banking, and venue altitude; road races vary in distance and conditions. Adjustment formulas exist in coaching and biomechanics literature~\cite{riegel1981athletic,maurer2018race,corts2017quantitative,gutierrez2025real} but have not been operationalized at scale in a single public collegiate corpus.

NRCD enables information-retrieval and data-mining research that access constraints have blocked: longitudinal athlete modeling across seasons, benchmarking nonlinear predictors~\cite{rigatti2017random,chen2015xgboost}, fairness analysis across environmental confounders, and reproducible sports-informatics benchmarks. The resource pairs a public dataset of NIRCA athletes, a validated conversion formula, a pip-installable `nrcd' Python package, and a live, moderated, contributed website on \url{https://www.nationalrunningclubdatabase.com}, addressing future sustainability. By supporting equitable participation research and evidence on training and competition under environmental stress, NRCD aligns with UN SDG 5 (Gender Equality) and UN SDG 3 (Good Health and Well-Being)~\cite{UN_SDG_TargetsOverview}.

\textbf{Contributions and scope:} This resource paper describes the \textbf{dataset} and a \textbf{standardization framework}, with empirical validation on cross country (XC): each fall meet is the same aerobic race for a given gender (8k/6k). Track mixes distances and event counts per athlete per meet, and road distances/conditions are inconsistent; track and road policies are specified but \emph{not} validated here. We do \emph{not} report predictive benchmarks here. Those belong in separate analytical papers. We present: \\

\noindent (1)~\textbf{NRCD} - 143,868 results (36.2\% women) across four sports; \\
(2)~\textbf{Community-governed curation} - results added by community and approved by domain experts;\\
(3)~\textbf{Environmentally aware standardization} - converted-only and standardized modes with a derived heat law ($k{=}0.0016$) and the \textbf{`nrcd' Python package};\\
(4)~\textbf{Gender-stratified empirical checks} - separate analyses for men and women.

\section{Related Work}
\textbf{Open running data:} Prior open collegiate work uses less than a thousand hand-curated results~\cite{millettpredicting}. No prior resource offers close to 100,000+ results, with course and weather metadata.

\textbf{Standardization.} Riegel's law~\cite{riegel1981athletic}, Maurer elevation factors~\cite{maurer2018race}, and weather--performance work~\cite{cusick2023impact,zender2024effect,MaxPerformanceRunning} exist separately; NRCD integrates them with a quadratic heat surrogate fit to Hadley's coaching bands~\cite{zhao2015effect}, excluding AQI adjustment~\cite{cusick2023impact}.

\textbf{Community sports data:} Fitness apps record training, not officiated meets~\cite{gute2022keep,mckenzie2017ised}. NRCD provides governed meet results with unique ids for user tracking.

\section{The NRCD Dataset}
\begin{figure*}
    \centering
    \includegraphics[width=.81\linewidth]{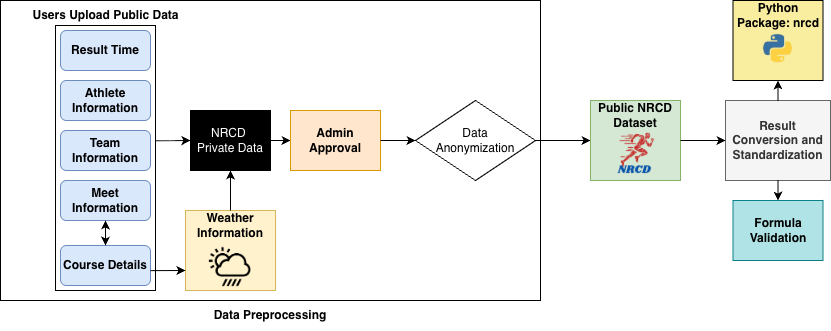}
    \caption{NRCD data and standardization pipeline}
    \label{fig:pipeline}
\end{figure*}
NRCD covers \emph{Cross Country}, \emph{Indoor Track}, \emph{Outdoor Track}, and \emph{Road Race}. Tables~\ref{tab:composition}-\ref{tab:coverage} summarize the approved export; Table~\ref{tab:coverage} splits \textbf{comprehensive} (meet date $\geq$ August 2023, richer metadata) from \textbf{historical} (2003-July 2023). Of 143,868 results, 65,622 are comprehensive and 78,246 historical. The same \texttt{athlete\_id} links sports; 11,601 athletes (37.0\%) appear in $\geq$2 school years and 5,584 (17.8\%) in $\geq$3.

\begin{table}[t]
  \centering
  \small
  \begin{tabular}{lrrr}
    \toprule
    Sport & Results & Athletes$^\dagger$ & Meets \\
    \midrule
    Cross Country & 75,489 & 23,565 & 546 \\
    Indoor Track    & 29,595 &  9,657 & 348 \\
    Outdoor Track   & 36,598 &  9,976 & 469 \\
    Road Race       &  2,186 &  1,630 &  60 \\
    \midrule
    \textbf{Total} & \textbf{143,868} & \textbf{31,351} & \textbf{1,423} \\
    \bottomrule
  \end{tabular}
  \caption{Results by sport}
  \label{tab:composition}
  \vspace{-0.4em}
  {\footnotesize $^\dagger$Each \emph{Athletes} cell counts distinct ids. Sport rows are not summable: 31.2\% of athletes are multi-sport.}
\end{table}

\begin{table}[t]
  \centering
  \footnotesize
  \setlength{\tabcolsep}{3pt}
  \resizebox{\columnwidth}{!}{%
  \begin{tabular}{llrrrrr}
    \toprule
    Sport & Era & Results & Athletes$^\dagger$ & Weather\ (\%) & Alt.\ (\%) & Course\ (\%) \\
    \midrule
    Cross Country & historical & 52,129 & 17,865 & 27.2 & 97.8 & 3.0 \\
    Cross Country & comprehensive & 23,360 &  7,056 & 99.9 & 100.0 & 38.5 \\
    Indoor Track    & historical & 11,386 &  5,118 & --- & 100.0 & --- \\
    Indoor Track    & comprehensive & 18,209 &  5,174 & --- & 99.4 & --- \\
    Outdoor Track   & historical & 13,399 &  5,098 & 0.0 & 100.0 & --- \\
    Outdoor Track   & comprehensive & 23,199 &  5,665 & 35.0 & 99.9 & --- \\
    Road Race       & historical &  1,332 &  1,047 & 93.4 & 100.0 & 52.9 \\
    Road Race       & comprehensive &   854 &    612 & 100.0 & 100.0 & 33.4 \\
    \bottomrule
  \end{tabular}%
  }
  \caption{Total coverage by sport and era (Alt. = altitude, Course = gain/loss and/or short/long course details)}
  \label{tab:coverage}
\end{table}

\begin{table}[t]
  \centering
  \small
  \begin{tabular}{lrrr}
    \toprule
    Sport & Women & Men & Women (\%) \\
    \midrule
    Cross Country & 29,113 & 46,376 & 38.6 \\
    Indoor Track    &  9,965 & 19,630 & 33.7 \\
    Outdoor Track   & 11,967 & 24,631 & 32.7 \\
    Road Race       &    986 &  1,200 & 45.1 \\
    \midrule
    \textbf{Total} & \textbf{52,031} & \textbf{91,837} & \textbf{36.2} \\
    \textbf{Unique athletes} & \textbf{12,364} & \textbf{18,987} & \textbf{39.4} \\
    \bottomrule
  \end{tabular}
  \caption{Results by sport and gender}
  \label{tab:gender}
\end{table}

\begin{table}[t]
  \centering
  \scriptsize
  \setlength{\tabcolsep}{1.5pt}
  \resizebox{\columnwidth}{!}{%
  \begin{tabular}{llrrrrrr}
    \toprule
    Sport & Gender & Sprint & Mid & Dist. & Relay & Field & Total \\
    \midrule
    Cross Country & F & 0 & 0 & 29,113 & 0 & 0 & 29,113 \\
    Cross Country & M & 0 & 0 & 46,376 & 0 & 0 & 46,376 \\
    Indoor Track    & F & 2,851 & 1,018 & 3,453 & 1,129 & 1,514 & 9,965 \\
    Indoor Track    & M & 6,184 & 2,019 & 7,175 & 2,003 & 2,249 & 19,630 \\
    Outdoor Track   & F & 3,853 & 1,563 & 3,904 & 729 & 1,918 & 11,967 \\
    Outdoor Track   & M & 7,808 & 3,425 & 8,977 & 1,534 & 2,887 & 24,631 \\
    Road Race       & F & 0 & 0 & 986 & 0 & 0 & 986 \\
    Road Race       & M & 0 & 0 & 1,200 & 0 & 0 & 1,200 \\
    \midrule
    \textbf{All} & - & \textbf{20,696} & \textbf{8,025} & \textbf{101,184} & \textbf{5,395} & \textbf{8,568} & \textbf{143,868} \\
    \bottomrule
  \end{tabular}%
  }
  \caption{Results by category. Sprint.: $\leq$400\,m and hurdles; mid: 401-1000\,m; dist.: $>$1000\,m and steeplechase.}
  \label{tab:events}
\end{table}

\section{Community-Governed Curation}
NRCD is \textbf{open-contributed}: coaches, athletes, and volunteers submit meets; no entry is added until an admin with domain expertise approves it against public postings. Admins also review public meet results and registered teams nationwide to add missing meets, so coverage is broader than user submissions alone. Approved meets grew from 78 to 145, 335, and 274 per year (2022-2025). Potential athlete merges follow a standard record-linkage procedure~\cite{christen2012data}: the system \emph{suggests} pairs that exceed 90\% similarity and an admin reviews each suggestion. Parallel queues handle potential meet merges and meet-series links across years. Public exports will continue to be updated \textbf{yearly}. Exports strip PII and include datasheets~\cite{wilkinson2016fair,gebru2021datasheets}; weather via OpenWeatherMap is included where course records exist~\cite{OpenWeatherMapAPI}.

\section{Unified Standardization Framework}
\label{sec:standardization}
Figure~\ref{fig:pipeline} summarizes the pipeline: approved results join athletes, teams, events, and course metadata; sport-specific factors (Table~\ref{tab:factors}) yield comparable times \emph{within} sport and gender. For XC, factors apply in fixed order: weather, grade, venue altitude ($f_{\mathrm{alt}}$), measured-vs.-reported course length, then Riegel conversion to $d_{\mathrm{target}}$ (Equation~\ref{eq:xc}; heat in Section~\ref{sec:heatvalid}).

Raw time $t_{\mathrm{raw}}$ maps to standardized time via multiplicative factors and Riegel conversion~\cite{riegel1981athletic}:
\begin{equation}
  t_{\mathrm{std}} = t_{\mathrm{raw}} \cdot \prod_i f_i \cdot \left(\frac{d_{\mathrm{target}}}{d_{\mathrm{actual}}}\right)^{b},
  \quad b \in \{1.055\ (\mathrm{M}),\ 1.08\ (\mathrm{F})\}.
  \label{eq:general}
\end{equation}
\footnote{We adopt Riegel's gender-specific fatigue factors~\cite{riegel1981athletic} ($b{=}1.055$ men, $1.08$ women), not a pooled population $b$. Classical Riegel and later predictors often use $b{=}1.06$~\cite{blythe2016prediction}; modern record refits vary by discipline ($\approx$1.04--1.12). On NRCD XC, substituting $b{=}1.06$ changes validation percentages by ${<}0.5$~pp.}
For XC,
\begin{equation}
  t_{\mathrm{std}} = t_{\mathrm{raw}} \cdot f_{\mathrm{weather}} \cdot f_{\mathrm{elev}} \cdot f_{\mathrm{alt}} \cdot \left(\frac{d_{\mathrm{actual}}}{d_{\mathrm{reported}}}\right)^{b} \cdot \left(\frac{d_{\mathrm{target}}}{d_{\mathrm{actual}}}\right)^{b},
  \label{eq:xc}
\end{equation}
with $d_{\mathrm{target}} \in \{8000\,\mathrm{m}, 6000\,\mathrm{m}\}$ for men and women (NIRCA defaults)~\cite{nirca}. The same exponent $b$ is used twice: $(d_{\mathrm{actual}}/d_{\mathrm{reported}})^{b}$ corrects short or long measured courses relative to the listed distance, and $(d_{\mathrm{target}}/d_{\mathrm{actual}})^{b}$ converts to the gender reference (8\,km/6\,km). 

\textbf{Gender-stratified analysis:} \textbf{We recommend fitting, evaluating, and reporting all models separately by gender} (Section~\ref{sec:validation}), never pooling raw or adjusted seconds across men and women.

\textbf{Temperature:} Heat index $H$ is temperature plus dew point ($^\circ$F)~\cite{MaxPerformanceRunning}. We fit a batchable quadratic surrogate to Hadley's coaching bands~\cite{zhao2015effect} (not peer-reviewed physiology):
\begin{equation}
  p(H) = k\,(H-100)^2,\quad
  f_{\mathrm{weather}} = 1 - \frac{p(H)}{100},\quad H>100,
  \label{eq:heat}
\end{equation}
with $f_{\mathrm{weather}}{=}1$ for $H{\leq}100$, for aerobic events per Table~\ref{tab:factors}. We fit $k{=}0.0016$ to Hadley midpoints only (Table~\ref{tab:heatfit}), not finish times.

\textbf{AQI:} Measured where available but omitted from $f_i$ since it is confounded with temperature and dew point~\cite{cusick2023impact}.

\textbf{Elevation and course length:} $f_{\mathrm{elev}} = 1.04^{g} \cdot 0.9633^{l}$ for gain/loss grades $g$, $l$~\cite{maurer2018race} (ft$\to$grade\% via measured length). Measured course length differing from reported distance adds $(d_{\mathrm{actual}}/d_{\mathrm{reported}})^{b}$ before converting to $d_{\mathrm{target}}$.

\textbf{Venue altitude:} $f_{\mathrm{alt}}=t_{\mathrm{sea}}/t_{\mathrm{alt}}$ from the gender-stratified P\'eronnet-Thibault model~\cite{peronnet1989mathematical,peronnet1991theoretical} at $d_{\mathrm{actual}}$: MAP reduced at meet elevation; air density from barometric pressure when recorded.

\textbf{Converted only vs.\ standardized:} We present \emph{converted} (distance and course-length, no weather/elevation) and \emph{standardized} adjustments. Converted-only times can absorb fall-season cooling and grade, inflating apparent within-season fitness gains (Section~\ref{sec:improvement}).

\textbf{Track:} Equation~\ref{eq:general} with NCAA facility-indexing multipliers~\cite{ncaa2012indoorconversion,ncaa2012facilityindex,barnes2017conversion,corts2017quantitative} (tabulated $\alpha$; repo lookup). Outdoor sprint/hurdle wind uses per-result \texttt{wind} with event-specific calm-equivalent $\Delta t$~\cite{linthorne1994effect,quinn2003effects,quinn2004effects,mureika2001legality,spiegel2003model}: 100~m (Linthorne); 200~m (Quinn Table~2, straight-gauge); 110~m hurdles (Spiegel \& Mureika); 400~m (Quinn's curved-track simulation); same $f_{\mathrm{alt}}$ when $z$ is set; heat on outdoor mid/distance.

\textbf{XC and Road:} Equation~\ref{eq:general} with Riegel, $f_{\mathrm{weather}}$, $f_{\mathrm{elev}}$, course-distance accuracy, and $f_{\mathrm{alt}}$ when metadata exist.

\begin{table}[t]
  \centering
  \footnotesize
  \setlength{\tabcolsep}{3pt}
  \resizebox{\columnwidth}{!}{%
  \begin{tabular}{lcccc}
    \toprule
    Factor & Cross Country & In.\ track & Out.\ track & Road \\
    \midrule
    Riegel distance & \checkmark & \checkmark & \checkmark & \checkmark \\
    Heat ($f_{\mathrm{weather}}$) & \checkmark & --- & mid+ & \checkmark \\
    Wind (per-result) & --- & --- & sprint & --- \\
    Elevation gain/loss & \checkmark & --- & --- & \checkmark \\
    Course distance accuracy & \checkmark & --- & --- & \checkmark \\
    AQI* & \checkmark & --- & \checkmark & \checkmark \\
    Altitude & \checkmark & \checkmark & \checkmark & \checkmark \\
    Track length \& banking  & --- & \checkmark & --- & --- \\
    \bottomrule
  \end{tabular}%
  }
  \caption{Adjustment factors by sport (AQI only measured)}
  \label{tab:factors}
\end{table}

\begin{table}[t]
  \centering
  \small
  \setlength{\tabcolsep}{3pt}
  \begin{tabular}{ccccc}
    \toprule
    $H$ & Hadley (\%) & Mid (\%) & Quad.\ (\%) & $f_{\mathrm{weather}}$ \\
    \midrule
    110 & 0.0-0.5 & 0.25 & 0.16 & 0.9984 \\
    120 & 0.5-1.0 & 0.75 & 0.64 & 0.9936 \\
    145 & 3.0-4.5 & 3.75 & 3.24 & 0.9676 \\
    160 & 4.5-6.0 & 5.25 & 5.76 & 0.9424 \\
    175 & 8.0-10.0 & 9.00 & 9.00 & 0.9100 \\
    \bottomrule
  \end{tabular}
  \caption{Hadley bands vs.\ quadratic surrogate ($k{=}0.0016$).}
  \label{tab:heatfit}
\end{table}

\section{Formula Validation (XC)}
\label{sec:validation}
We validate on XC because athletes typically race one distance across a season (women 6\,km, men 8\,km), enabling within-athlete cross-meet checks. Track mixes events and road race regularity is inconsistent, so they are \emph{not} validated here. All summaries are \textbf{gender-stratified} for XC; never pooled together.

\subsection{Heat Surrogate}
\label{sec:heatvalid}
\textbf{(1)~Band fidelity:} $k{=}0.0016$ matches Hadley slowdown ranges in percent (Table~\ref{tab:heatfit}), with RMSE 0.34~pp vs.\ band midpoints and 70\% of $H\in[101,180]$ inside published ranges (fidelity to the heuristic~\cite{MaxPerformanceRunning}, not independent physiological validation). \textbf{(2)~Structure:} $f_{\mathrm{weather}}{=}1$ for $H{\leq}100$; monotone decrease as $H$ rises. \textbf{(3)~Plausibility:} on 57,802 XC rows with weather, mean factor $0.990$ when $H{>}100$. Finish-time recalibration of $k$ is unstable and rejected. Validation metrics are stable for $k \in \{0.001, 0.0016, 0.002\}$ (Table~\ref{tab:ksens}).

\begin{table}[t]
  \centering
  \footnotesize
  \setlength{\tabcolsep}{3pt}
  \begin{tabular}{lrrrr}
    \toprule
    $k$ & W reduc.\ (\%) & M reduc.\ (\%) & W infl.\ (\%) & M infl.\ (\%) \\
    \midrule
    0.0010 & 50.9 & 35.1 & 9 & 13 \\
    0.0016 & 51.1 & 35.4 & 15 & 20 \\
    0.0020 & 50.9 & 35.1 & 19 & 24 \\
    \bottomrule
  \end{tabular}
  \caption{XC heat-$k$ sensitivity. Reduc.\ = cross-meet SD; inflat.\ = converted vs.\ full.}
  \label{tab:ksens}
\end{table}

\subsection{Within-Athlete Variance}
For athletes with $\geq 3$ regular-season XC meets per season (nationals excluded), Table~\ref{tab:variance} reports median cross-meet SD on raw, converted-only, and standardized times (6\,km/8\,km). Reductions vs.\ raw under standardization (51.1\% women, 35.4\% men) are mostly mechanical distance conversion (96.7\%/87.3\% of the SD drop); converted to standardized isolates the environmental stack (3.3\%/6.5\% further SD reduction). Lower variance is necessary but not sufficient; Section~\ref{sec:improvement} tests whether environment removes confounding.

\begin{table}[t]
  \centering
  \small
  \begin{tabular}{lrrrrr}
    \toprule
    Group & $n$ & Raw & Conv. & Std. & Reduct. \\
    \midrule
    Women (6\,km)  & 2,201 & 108.09 & 54.66 & 52.86 & 51.1\% \\
    Men (8\,km)    & 3,984 & 85.56 & 59.12 & 55.27 & 35.4\% \\
    \bottomrule
  \end{tabular}
  \caption{Median within-athlete cross-meet SD (XC, $\geq 3$ meets/season). Reduct.\ = raw vs.\ standardized.}
  \label{tab:variance}
\end{table}

\subsection{Within-Season Improvement}
\label{sec:improvement}
For athletes with $\geq 2$ regular-season races, \emph{improvement} is first-race minus best later finish times on the gender reference distance (positive = faster later). Table~\ref{tab:improvement} compares \emph{converted only} versus \emph{standardization} on the same athlete-seasons; brackets give 95\% percentile-bootstrap CIs (2,000 replicates).

\begin{table}[t]
  \centering
  \footnotesize
  \setlength{\tabcolsep}{2pt}
  \begin{tabular}{lrlrlr}
    \toprule
    Group & $n$ & Conv.\ only & Full std. & Inflat.\ (\%) \\
    \midrule
    Women & 5,653 & 44.4 {\scriptsize[41.7, 47.0]} & 38.6 {\scriptsize[35.5, 41.5]} & 15 {\scriptsize[11, 21]} \\
    Men   & 9,130 & 41.0 {\scriptsize[39.0, 43.3]} & 34.2 {\scriptsize[32.4, 36.5]} & 20 {\scriptsize[17, 24]} \\
    \bottomrule
  \end{tabular}
  \caption{Median within-season improvement (sec) with 95\% bootstrap CIs. Inflat.\ (\%) $= 100\times(\mathrm{median}_{\mathrm{conv}}/\mathrm{median}_{\mathrm{std}} - 1)$.}
  \label{tab:improvement}
\end{table}

Converted-only inflates median within-season improvement by 15\% (women) and 20\% (men) versus full standardization because fall cooling and grade remain in converted-only times. Gender-stratified RF/GB (train 2023, test 2024) predict improvement rate (s/day) from season-level XC features: converted-only times often meet or exceed standardized test $R^2$ (RF men 0.740 vs.\ 0.728, women 0.712 vs.\ 0.715; GB men 0.587 vs.\ 0.561, women 0.592 vs.\ 0.516; only GB women significant). The paired $\Delta R^2$ is an external check the variance drop alone cannot give; environment in non-standardized times is residual confounding. \textbf{Without full standardization, models risk predicting external factors, not the athlete.}

\textbf{Limitations:} Statistics only reflect NIRCA club competition. Pre-August 2023 entries lack complete metadata. Results are adjusted for environmentally aware standards, not individual training plans. Variance and improvement validation only uses XC.

\section{Conclusion}
NRCD is the first open large-scale collegiate running dataset with documented standardization. A derived heat surrogate is validated against Hadley's piecewise bands, not finish-time regression. Gender-stratified XC checks show that full standardization reduces within-athlete cross-meet variability (51.1\% women, 35.4\% men vs.\ raw) and avoids overstating within-season improvement relative to converted-only times. We recommend gender-stratified modeling on all derivative work. Finally, we release a Python standardization package `nrcd', removing the bulk-export barrier for sports informatics.

\section*{Ethics Statement}
All of our data is obtained from the NRCD, which is publicly accessible. We remove all PII such as links and persons' names and have research permission to use and publish NRCD's dataset. The University of Notre Dame IRB classified this work as not human subjects research (protocol~25-05-9348). Users must not attempt to re-identify athletes or merge NRCD with external identity sources.

\begin{acks}
We greatly appreciate Tom Meurer for supporting the initial project in Notre Dame’s Advanced Database class. We also thank the members and staff of NIRCA, the CRC Coaches Poll, the Notre Dame Running Club, and  Notre Dame RecSports for their support.
\end{acks}

\section*{Generative AI Usage Disclosure}
Moderate AI use (Cursor, Claude) was used for code generation and paper editing, though all content is validated as our own. No AI was used for result upload or validation.

%% file: appendix.tex

\section*{Appendix}

\section{Supplementary Validation}
\label{app:validation}

\subsection{Stepwise variance decomposition (XC)}
\label{app:variance-steps}

For athlete-seasons with $\geq 3$ regular-season XC meets (nationals excluded), Table~\ref{tab:app-variance-steps} reports median cross-meet standard deviation for four result-time variants: \emph{raw}; \emph{converted-only} (Riegel + measured course length); \emph{pre-weather env.} (grade, altitude, course length on top of converted); and \emph{fully standardized} (+ heat). Raw$\to$full is percent SD reduction from raw to the final column.

\begin{table}[t]
  \centering
  \footnotesize
  \setlength{\tabcolsep}{3pt}
  \resizebox{\columnwidth}{!}{%
  \begin{tabular}{lrrrrrr}
    \toprule
    Group & $n$ & Raw & Conv. & Pre-wx & Full & Raw$\to$full \\
    \midrule
    Women (6\,km) & 2,201 & 108.1 & 54.7 & 54.6 & 52.9 & 51.1\% \\
    Men (8\,km)   & 3,984 & 85.6 & 59.1 & 58.7 & 55.3 & 35.4\% \\
    \bottomrule
  \end{tabular}%
  }
  \caption{Median within-athlete cross-meet SD (seconds) by adjustment stage.}
  \label{tab:app-variance-steps}
\end{table}

\textbf{What drives the gender gap?} Most reduction occurs at the \emph{distance-conversion} step (women: raw 108\,s to converted 55\,s). Of the raw$\to$full median-SD drop, distance conversion accounts for 96.7\% (women) and 87.3\% (men); the environmental stack (grade, altitude, heat) accounts for the remaining 3.3\% and 12.7\%. Equivalent headline numbers isolating environment: converted-only $\to$ full standardization further reduces median cross-meet SD by 3.3\% (women) and 6.5\% (men). Pre-weather columns (grade + altitude + length, no heat) move medians only ${\sim}$0.1 to 0.4\,s; heat adds another ${\sim}$1.7 to 3.4\,s. Women's larger raw$\to$full percent (51.1\% vs.\ 35.4\%) therefore reflects a wider pre-adjustment distribution on heterogeneous course distances, not larger heat-index swings. Distance conversion \emph{should} dominate SD when listed distances vary; lower raw$\to$full variance is necessary but not sufficient for the environmental stack. That stack is independently visible as 15\% / 20\% within-season improvement inflation (main text Table~\ref{tab:improvement}) and paired $\Delta R^2$ (Table~\ref{tab:app-ml-r2}).

\begin{table}[t]
  \centering
  \footnotesize
  \setlength{\tabcolsep}{2pt}
  \resizebox{\columnwidth}{!}{%
  \begin{tabular}{lrrrr}
    \toprule
    Group & Dist.\ share & Env.\ share & Conv$\to$full & Infl.\ (\%) \\
    \midrule
    Women & 96.7\% & 3.3\% & 3.3\% & 15 \\
    Men   & 87.3\% & 12.7\% & 6.5\% & 20 \\
    \bottomrule
  \end{tabular}%
  }
  \caption{Attribution of XC raw$\to$full median-SD drop (Dist./Env.\ shares) vs.\ converted$\to$full SD reduction and improvement inflation.}
  \label{tab:app-env-attribution}
\end{table}

\subsection{Environmental factor distributions (comprehensive XC)}
\label{app:factor-dist}

Table~\ref{tab:app-factor-dist} summarizes adjustment factors on comprehensive-era XC results ($n{=}20{,}190$ after nationals exclusion). Medians near 1.0 confirm that population-level corrections are small on average; tails matter for individual meets.

\begin{table}[t]
  \centering
  \footnotesize
  \setlength{\tabcolsep}{3pt}
  \begin{tabular}{lrrrr}
    \toprule
    Factor subset & Median & Mean & P05 to P95 & \% outside $[0.95,1.05]$ \\
    \midrule
    Weather (all rows)        & 0.993 & 0.990 & 0.967 to 1.000 & 0.5 \\
    Weather ($H{>}100$ only)   & 0.990 & 0.988 & 0.966 to 1.000 & 0.6 \\
    Elevation (grade)         & 1.000 & 1.000 & 1.000 to 1.002 & 0.0 \\
    Altitude (P\'eronnet)       & 0.996 & 0.996 & 0.993 to 1.000 & 0.0 \\
    \bottomrule
  \end{tabular}
  \caption{Factor $f$ such that adjusted time $= f \times$ prior result time (comprehensive XC).}
  \label{tab:app-factor-dist}
\end{table}

\textbf{Elevation check.} After converting stored gain/loss from feet to grade percent before Maurer factors, the elevation median is exactly 1.0 with P95 1.002, consistent with the corrected unit convention. Weather drives the only material tail mass among active factors at the median athlete-season.

\subsection{Within-season improvement direction}
\label{app:improvement-direction}

Converted-only result times overstate not only \emph{magnitude} of within-season improvement (15\% / 20\% median inflation for women / men) but also the \emph{share} of athlete-seasons classified as improving (first race slower than later best): 69.8\% / 68.1\% under converted-only vs.\ 67.6\% / 66.2\% under full standardization (${\sim}$1.9 to 2.2\,pp gap). Directional bias is modest but aligned with inflation: environmental drift in unadjusted result times mimics fitness gains.

\subsection{Illustrative ML $R^2$ and bootstrap uncertainty}
\label{app:ml-r2}

The illustrative analysis is a \emph{demonstration}, not a benchmark: gender-stratified models predict within-season improvement \emph{rate} (s/day) from season-level features derived from the same result times (train 2023, test 2024; $\geq 2$ XC meets). Table~\ref{tab:app-ml-r2} reports random forest and gradient boosting; 95\% CIs are percentile bootstrap on the test set (2,000 replicates).

\begin{table}[t]
  \centering
  \footnotesize
  \setlength{\tabcolsep}{3pt}
  \begin{tabular}{llrrl}
    \toprule
    Model & Gender & Conv. & Std. & $\Delta R^2$ 95\% CI \\
    \midrule
    RF & Men   & 0.740 & 0.728 & $[-0.006,\ +0.037]$ \\
    RF & Women & 0.712 & 0.715 & $[-0.034,\ +0.034]$ \\
    GB & Men   & 0.587 & 0.561 & $[-0.006,\ +0.059]$ \\
    GB & Women & 0.592 & 0.516 & $[+0.033,\ +0.125]$ \\
    \bottomrule
  \end{tabular}
  \caption{Test $R^2$ by result-time adjustment. $\Delta R^2 = R^2_{\mathrm{conv}} - R^2_{\mathrm{std}}$; 95\% CI excludes 0 only for GB women.}
  \label{tab:app-ml-r2}
\end{table}

\textbf{Why report $R^2$ at all?} Features include within-season time summaries ($\mathrm{first\_time}$, $\mathrm{best\_time}$, $\mathrm{slope}$, etc.) built from the result times under test, so high $R^2$ is expected and is \emph{not} external predictive validity. The relevant comparison is \emph{paired}: same features, same splits, two result-time variants. Converted-only times embed fall cooling and course grade; standardized times partial those out. Higher $R^2$ under converted-only therefore signals that the model can exploit environmental structure left in the label (confounding, not added athlete signal).

\textbf{Why RF $\approx$ GB?} Direction is usually converted $\geq$ standardized (RF women is essentially tied within bootstrap noise). RF gaps are small and not significant at 95\%; GB gaps are larger, with women's GB $\Delta R^2$ significant (CI excludes 0). This suggests that a flexible but regularized fit still exploits environmental structure left in converted-only labels more than RF does. Neither model uses weather columns directly; environment enters only through result times.

\subsection{Extended robustness checks}
\label{app:robustness}

\subsubsection{Riegel exponent sensitivity}
\label{app:riegel-sens}

Gender-specific exponents ($b{=}1.055$ men, $b{=}1.08$ women) are the \emph{default design choice} from Riegel's table, not a post-hoc fit. A unified $b{=}1.06$ (classical single-exponent simplification~\cite{blythe2016prediction}) shifts XC variance-validation percentages by at most $0.4$~pp (Table~\ref{tab:app-riegel-sens}). The larger $1.0$~pp shift appears only under the stress test that substitutes $b{=}1.08$ for both genders (men's reduction 35.4\%$\to$36.4\%), which is not our deployed configuration.

\begin{table}[t]
  \centering
  \footnotesize
  \setlength{\tabcolsep}{2pt}
  \resizebox{\columnwidth}{!}{%
  \begin{tabular}{lrrr}
    \toprule
    Riegel config & Men raw$\to$std & Women raw$\to$std & Max $|\Delta|$pp \\
    \midrule
    Gender-specific (default) & 35.4\% & 51.1\% & 0.0 \\
    Unified $b{=}1.06$          & 35.8\% & 50.9\% & 0.4 \\
    Unified $b{=}1.055$         & 35.4\% & 50.8\% & 0.3 \\
    Unified $b{=}1.08$          & 36.4\% & 51.1\% & 1.0 \\
    \bottomrule
  \end{tabular}%
  }
  \caption{Median within-athlete SD reduction under alternate Riegel exponents ($\geq 3$ meets/season).}
  \label{tab:app-riegel-sens}
\end{table}

\subsubsection{Finisher-only export and minimum-race sensitivity}
\label{app:finisher-sens}

The public export contains approved \emph{finisher} times only; DNS/DNF are absent as rows. XC improvement inflation is stable when requiring $\geq 2$, $\geq 3$, or $\geq 4$ races per athlete-season (men 16.8 to 19.9\%, women 7.5 to 15.0\%; Table~\ref{tab:app-finisher-sens}). Only 18.6\% of XC athlete-seasons have $\geq 3$ races, so sparse-season athletes dominate the $\geq 2$ race demo.

\begin{table}[t]
  \centering
  \small
  \begin{tabular}{lrrrr}
    \toprule
    Min races & Men $n$ & Men infl.\ & Women $n$ & Women infl.\ \\
    \midrule
    2 & 9,130 & 19.9\% & 5,653 & 15.0\% \\
    3 & 4,055 & 18.9\% & 2,233 & 7.5\% \\
    4 & 1,340 & 16.8\% & 739 & 13.5\% \\
    \bottomrule
  \end{tabular}
  \caption{Converted-only median improvement inflation vs.\ full standardization by minimum races per athlete-season.}
  \label{tab:app-finisher-sens}
\end{table}

\subsubsection{Outdoor track wind and venue factors}
\label{app:track-wind}

Comprehensive outdoor track ($n{=}20{,}028$ parseable results): sprint/hurdle events ($\leq 400$\,m) with wind logic applicable on 7,429 rows, but wind \emph{recorded} on only 7.0\% of those. Among applicable rows the wind-factor median is 1.0 (P95 1.055). NCAA venue length/banking factors are unity at the median for this corpus (lap metadata sparse). The full track pipeline (wind when present, heat, altitude) yields median factor 0.999 with 9.8\% of rows outside $[0.98, 1.02]$; meet-level weather coverage remains 4.6\%.

\subsubsection{Cross-sport standardization magnitude}
\label{app:cross-sport}

Table~\ref{tab:app-cross-sport} summarizes comprehensive-era adjustment factors (adjusted/raw) for running events. XC shows the largest tail mass (52.4\% outside $[0.98, 1.02]$) because distance conversion and course metadata move times materially. Indoor track is tight at the median (0.09\% outside band); outdoor track and road sit between.

\begin{table}[t]
  \centering
  \small
  \begin{tabular}{lrrr}
    \toprule
    Sport & $n$ & Median factor & \% outside $[0.98, 1.02]$ \\
    \midrule
    Cross Country   & 31,450 & 0.985 & 52.4 \\
    Outdoor Track   & 18,619 & 0.998 & 9.7 \\
    Indoor Track    & 14,693 & 1.000 & 0.1 \\
    Road Race       &    854 & 0.995 & 2.3 \\
    \bottomrule
  \end{tabular}
  \caption{Comprehensive-era running results: full sport-appropriate adjustment pipeline.}
  \label{tab:app-cross-sport}
\end{table}

\subsubsection{Road race exploratory transfer checks}
\label{app:road-validation}

Road racing shares XC's Riegel + weather + elevation stack (main-text Table~\ref{tab:factors}) but does \emph{not} collapse every finish to a single gender target distance: a 5\,km and a half marathon remain different events after length correction. Within-athlete SD on mixed road distances is therefore dominated by event choice, not environment, and XC-style variance reduction is the wrong primary metric. We instead report (i)~comprehensive-era factor magnitudes and (ii)~rank correlation between each athlete's best fully standardized XC time and best standardized road time as a transfer check that the shared stack preserves athlete ordering across surfaces.

\begin{table}[t]
  \centering
  \small
  \begin{tabular}{lrrr}
    \toprule
    Gender & $n$ both & Spearman $r$ & Med.\ XC best \\
    \midrule
    Women & 598 & 0.538 & 26:13 \\
    Men   & 688 & 0.431 & 29:05 \\
    \bottomrule
  \end{tabular}
  \caption{Athletes with $\geq 1$ XC and $\geq 1$ road result: Spearman correlation of career-best standardized times (XC median shown; road distances are heterogeneous so road medians are not comparable).}
  \label{tab:app-road-xc-corr}
\end{table}

On comprehensive road rows ($n{=}865$), median weather and elevation factors are near $1.0$ (weather mean $0.994$; 3.6\% of weather factors outside $[0.98,1.02]$). Among athletes with $\geq 2$ road finishes, environmental adjustment trims median within-athlete SD by only 0.2--1.5\%---consistent with sparse grade/weather metadata and heterogeneous race distances, not a failed pipeline. Moderate XC$\leftrightarrow$road Spearman correlations (Table~\ref{tab:app-road-xc-corr}) support transfer of the shared stack without claiming XC-grade validation for road. Indoor/outdoor track remain formula-shipped with NCAA/wind literature factors pending denser metadata and event-homogeneous validation designs.

\subsubsection{Meet-redacted temporal ML split}
\label{app:meet-holdout}

The illustrative ML demo uses train 2023 / test 2024. To stress meet-level leakage, we rebuild 2024 test features after randomly withholding 25\% of 2024 meets (25 meets) from each athlete's race history; labels are unchanged. Random-forest $R^2$ shifts slightly but the conv.\ vs.\ std.\ gap remains modest (men $\Delta R^2 {=} 0.000$; women $+0.044$; Table~\ref{tab:app-meet-holdout}), comparable in magnitude to the temporal bootstrap CIs in Table~\ref{tab:app-ml-r2}.

\begin{table}[t]
  \centering
  \small
  \begin{tabular}{lrrr}
    \toprule
    Gender & $R^2$ conv. & $R^2$ std. & $\Delta R^2$ \\
    \midrule
    Men   & 0.720 & 0.720 & $0.000$ \\
    Women & 0.771 & 0.727 & $+0.044$ \\
    \bottomrule
  \end{tabular}
  \caption{Meet-redacted temporal split (train 2023; test 2024 with 25\% of meets removed from feature construction).}
  \label{tab:app-meet-holdout}
\end{table}

\section{Cross Country Performance Band and NCAA Context}
\label{app:xc-percentiles}

NIRCA club competition spans NCAA Division~I-caliber athletes to recreational 5K-equivalent efforts. To contextualize ability spread, we report each athlete's career \emph{personal record} (PR): the fastest \textbf{fully standardized} XC result time across all school years and meets in the export, \textbf{including the NIRCA nationals championship} (8\,km men, 6\,km women). Tables~\ref{tab:app-xc-ncaa}--\ref{tab:app-xc-overlap} use these standardized PRs, not raw meet race times. Nationals are excluded from formula-validation summaries in the main text because they are a single end-of-season championship, but they belong in PR tables because many athletes peak there. One row per athlete, not per result or per season. Table~\ref{tab:app-xc-percentiles} summarizes the full export; Table~\ref{tab:app-xc-cohorts} restricts to \textbf{longitudinal subsets} used in season-trajectory and improvement work ($\geq 2$ or $\geq 3$ distinct school years with XC results; optional $\geq 2$ races in each qualifying year to mirror the main-text improvement demo, which still excludes nationals from within-season slopes). Multi-season athletes are faster at every percentile (retention bias: committed runners return; one-meet newcomers drop out).

\textbf{Anchor definition.} Collegiate comparison times are \emph{team scoring depth}, not individual place at nationals or whole-roster medians. XC teams score their top five finishers; sixth and seventh runners displace opponents but do not add points. We therefore anchor on the \textbf{5th runner} among the \textbf{top-15 teams} at each division's national championship---the slowest counting scorer on a top-tier team. This requires only five finishers per team (nearly all championship teams field at least five; only ${\sim}$95\% field seven). NIRCA anchors are fit from standardized NRCD nationals results (2015+); NCAA, NAIA, and NJCAA anchors are approximate from published championship team results. Anchors are illustrative, not formal equivalence tests.

\begin{table}[t]
  \centering
  \scriptsize
  \setlength{\tabcolsep}{1.5pt}
  \resizebox{\columnwidth}{!}{%
  \begin{tabular}{lrrrrrrrrrr}
    \toprule
    Gender & $n$ & P1 & P5 & P10 & P25 & P50 & P75 & P90 & P95 & P99 \\
    \midrule
    Men (8\,km)   & 13,962 & 25:03 & 26:09 & 26:49 & 28:16 & 30:20 & 33:02 & 36:16 & 38:54 & 47:02 \\
    Women (6\,km) &  9,603 & 21:53 & 23:07 & 23:52 & 25:21 & 27:19 & 29:43 & 32:32 & 34:30 & 40:13 \\
    \bottomrule
  \end{tabular}%
  }
  \caption{Career PR for \emph{all} athletes with $\geq 1$ parseable XC result (mm:ss); nationals included.}
  \label{tab:app-xc-percentiles}
\end{table}

\begin{table}[t]
  \centering
  \scriptsize
  \setlength{\tabcolsep}{1.5pt}
  \resizebox{\columnwidth}{!}{%
  \begin{tabular}{llrrrrrr}
    \toprule
    Cohort & Gender & $n$ & P5 & P50 & P95 \\
    \midrule
    All athletes        & Men   & 13,962 & 26:09 & 30:20 & 38:54 \\
    All athletes        & Women &  9,603 & 23:07 & 27:19 & 34:30 \\
    \addlinespace
    $\geq 2$ school years & Men   &  5,336 & 25:35 & 28:54 & 34:54 \\
    $\geq 2$ school years & Women &  3,321 & 22:30 & 25:54 & 31:16 \\
    $\geq 3$ school years & Men   &  2,596 & 25:19 & 28:16 & 33:27 \\
    $\geq 3$ school years & Women &  1,479 & 22:10 & 25:15 & 30:04 \\
    \addlinespace
    $\geq 2$ yr, $\geq 2$ races/yr & Men   &  3,224 & 25:22 & 28:24 & 33:40 \\
    $\geq 2$ yr, $\geq 2$ races/yr & Women &  1,930 & 22:11 & 25:22 & 30:04 \\
    \bottomrule
  \end{tabular}%
  }
  \caption{Career PR by longitudinal cohort (Aug~1 school-year boundary; Section~\ref{app:longitudinal}). Anchor exceedance by cohort is in Table~\ref{tab:app-xc-ncaa}.}
  \label{tab:app-xc-cohorts}
\end{table}

\begin{table}[t]
  \centering
  \footnotesize
  \setlength{\tabcolsep}{3pt}
  \begin{tabular}{lrr}
    \toprule
    Benchmark & Men (8k) & Women (6k) \\
    \midrule
    NCAA D1 All-American & ${\sim}$24:00 & ${\sim}$20:00 \\
    NCAA D1 5th runner (top-15 teams) & ${\sim}$26:00 & ${\sim}$22:00 \\
    NIRCA nationals 5th runner (NRCD fit) & ${\sim}$26:30 & ${\sim}$23:30 \\
    NCAA D2 5th runner & ${\sim}$28:00 & ${\sim}$24:00 \\
    NAIA 5th runner & ${\sim}$28:30 & ${\sim}$24:30 \\
    NCAA D3 5th runner & ${\sim}$29:30 & ${\sim}$25:30 \\
    NJCAA 5th runner & ${\sim}$28:30 & ${\sim}$25:30 \\
    \bottomrule
  \end{tabular}
  \caption{Standardized anchor \emph{times} (fastest$\to$slowest). ``5th runner'' $=$ fifth counting finisher on each of the top-15 teams at that division's nationals (Section~\ref{app:xc-percentiles}).}
  \label{tab:app-xc-anchor-times}
\end{table}

\begin{table}[t]
  \centering
  \footnotesize
  \setlength{\tabcolsep}{3pt}
  \begin{tabular}{lrr}
    \toprule
    Anchor$^\S$ & Men & Women \\
    \midrule
    NCAA D1 All-American & 0.1\% & 0.0\% \\
    NCAA D1 5th runner & 4.2\% & 1.3\% \\
    NCAA D2 5th runner & 21.6\% & 11.0\% \\
    NCAA D3 5th runner & 40.2\% & 26.9\% \\
    NAIA 5th runner & 27.7\% & 15.4\% \\
    NJCAA 5th runner & 27.7\% & 26.9\% \\
    \addlinespace
    NIRCA nationals 5th runner & 7.5\% & 7.3\% \\
    \bottomrule
  \end{tabular}
  \caption{Share of athletes whose standardized career PR is at or faster than each anchor (Table~\ref{tab:app-xc-anchor-times}). $^\S$Threshold labels shortened; times in companion table.}
  \label{tab:app-xc-ncaa}
\end{table}

\begin{table}[t]
  \centering
  \scriptsize
  \setlength{\tabcolsep}{2pt}
  \resizebox{\columnwidth}{!}{%
  \begin{tabular}{lrrl}
    \toprule
    Band (faster anchor $\to$ slower) & Men & Women & Interpretation \\
    \midrule
    D1 5th $\to$ NIRCA 5th & 3.3\% & 6.0\% & NIRCA depth, not D1 \\
    NIRCA 5th $\to$ D2 5th & 14.1\% & 3.7\% & D2 depth, not NIRCA \\
    NIRCA 5th $\to$ D3 5th & 32.7\% & 19.6\% & D3 depth, not NIRCA \\
    NIRCA 5th $\to$ NAIA 5th & 20.2\% & 8.1\% & NAIA depth, not NIRCA \\
    NIRCA 5th $\to$ NJCAA 5th & 20.2\% & 19.6\% & NJCAA depth, not NIRCA \\
    \bottomrule
  \end{tabular}%
  }
  \caption{Athletes between adjacent anchors (standardized career PR). Read row as: faster than the \emph{slower} anchor but slower than the \emph{faster} one.}
  \label{tab:app-xc-overlap}
\end{table}

\textbf{NIRCA vs.\ collegiate team depth.} Table~\ref{tab:app-xc-anchor-times} orders benchmarks; Table~\ref{tab:app-xc-ncaa} counts exceedance; Table~\ref{tab:app-xc-overlap} lists the \emph{between-anchor} bands that explain where NIRCA club runners sit relative to varsity nationals depth. For men, NIRCA (${\sim}$26:30) sits between D1 (${\sim}$26:00) and D2 (${\sim}$28:00); 3.3\% of athletes are in the NIRCA--D1 gap. For women, NIRCA (${\sim}$23:30) sits between D1 (${\sim}$22:00) and D2 (${\sim}$24:00): only 1.3\% of career PRs reach D1 5th-runner pace vs.\ 7.3\% at NIRCA and 11.0\% at D2.

\textbf{Interpretation.} About 25\% of career PRs occur at NIRCA nationals meets, so excluding them materially understated the elite tail. The full export (Table~\ref{tab:app-xc-percentiles}) still mixes one-meet recreational starters with multi-year club runners; P99 (men 47:02, women 40:13) reflects that long tail. Longitudinal cohorts (Table~\ref{tab:app-xc-cohorts}) are the relevant ability reference for papers that filter to $\geq 2$ school years (36.7\% of XC athletes in Section~\ref{app:longitudinal}) or require $\geq 2$ regular-season races per season before estimating improvement: median PR shifts ${\sim}$1:20 to 1:30 faster, and the share at NIRCA 5th-runner pace roughly doubles (e.g.\ men 7.5\% to 14--20\%; women 7.3\% to 14--20\%). Even in those subsets, NCAA D1 5th-runner pace remains uncommon ($\leq$12\% men, $\leq$4\% women).

\section{Metadata Missingness: Results vs.\ Meets}
\label{app:missingness}

Result-level coverage counts every row with a join to course metadata; meet-level coverage counts a meet if \emph{any} result in that meet has the field. Meet-level rates are lower when large invitationals contribute many results from a single weather-enriched row, and are often the more honest unit for season-trajectory analysis (one environment per competition day).

\begin{table}[t]
  \centering
  \footnotesize
  \setlength{\tabcolsep}{2pt}
  \resizebox{\columnwidth}{!}{%
  \begin{tabular}{llrrrr}
    \toprule
    Sport & Era & \multicolumn{2}{c}{Weather (\%)} & \multicolumn{2}{c}{Course feat.\ (\%)} \\
    & & Res. & Meet & Res. & Meet \\
    \midrule
    XC & comprehensive & 99.9 & 99.3 & 38.5 & 18.2 \\
    XC & historical     & 27.2 & 14.3 &  3.0 &  0.8 \\
    Out.\ track & comprehensive & 35.0 &  4.6 & ---& ---\\
    Road & comprehensive & 100.0 & 100.0 & 33.4 & 43.2 \\
    \bottomrule
  \end{tabular}%
  }
  \caption{Metadata coverage: share of results vs.\ share of meets (altitude omitted; ${>}97\%$ throughout).}
  \label{tab:app-missingness}
\end{table}

\textbf{Big-meet skew.} Among comprehensive-era XC meets ($n{=}280$), the largest 10 meets contain 28.0\% of all results; the largest 25 contain 45.5\%. Median meet size is 25 results (max 1,361). Result-level weather coverage therefore overstates how uniformly metadata blanket the calendar.

\textbf{Future releases.} Planned extensions include backfilling course gain/loss and measured length for high-volume invitationals, expanding outdoor-track weather joins, and documenting per-meet completeness in the datasheet. Historical rows (pre-August 2023) will remain sparse by design.

\subsection{Geolocation, altitude, and weather provenance}
\label{app:metadata-provenance}

Comprehensive-era metadata is attached to meets and courses, not to individual athletes or GPS tracks. Enrichment looks up the meet city and state through OpenWeather Geocoding and uses the first US match. One point represents the whole meet or course row, not points along the race route.

Venue altitude on the meet and course records is in feet from USGS EPQS at those coordinates. It reflects the geocoded host location (often near city center), not elevation sampled along the course. A rural park can be several miles from that point.

Weather comes from OpenWeather One Call 3.0 timemachine at the same coordinates. The listed race date and start time are converted to local time (via TimeZoneDB, or the meet timezone when stored). OpenWeather returns one hourly snapshot, not a minute-level reading. For example, a 10:55\,am start uses the 10:00\,am hour block. Conditions at the API point may differ from a wooded or exposed part of the course. Air-quality history is only available for meets on or after 27 November 2020. Indoor track is not weather-enriched. See Table~\ref{tab:app-missingness} for coverage rates and Section~\ref{app:uncovered-pacing} for other race-context factors standardization does not capture.

\subsection{Potential athlete, meet, and series merges}
\label{app:dedup}

Duplicate athlete profiles arise when the same runner is entered under slightly different spellings across meets (e.g., ``Jonathan'' vs.\ ``Jon'', truncated forms, middle initials present/absent). The live site does \emph{not} auto-merge. Admin queues follow a standard record-linkage pipeline~\cite{christen2012data}: preprocess, block, compare, then classify accept/reject. Rejecting a pair hides it from future suggestions. False merges are treated as higher cost than residual duplicates, so borderline cases are typically left unmerged. The public export therefore may still contain a small number of split identities; researchers studying career trajectories should treat multi-id athletes as a residual linkage error rather than assuming perfect entity resolution.

\subsubsection{Potential athlete merges}
\label{app:athlete-merges}

\begin{enumerate}
  \item \textbf{Preprocess} names (case/punctuation folded; nickname and initial forms retained for comparison).
  \item \textbf{Block} on same team $+$ exact last name $+$ first initial (e.g., John Smith and Johnny Smith).
  \item \textbf{Compare} given names with nickname canonization, truncated/initial prefix agreement (e.g., A$\leftrightarrow$Antonio, Al$\leftrightarrow$Alexander), or Jaro--Winkler $\geq 0.90$.
  \item \textbf{Classify} accept/reject in the admin UI.
\end{enumerate}

Same-gender prefix matches offer one \emph{Keep longer name} action. Other same-gender matches offer \emph{Choose~1} / \emph{Choose~2}. Opposite-gender pairs, ambiguous blocks ($3+$ athletes sharing the same last name $+$ first initial on a team), and pairs whose result date ranges do not overlap and are more than a year apart are listed for review but have no quick approve---admins use \emph{Open merge} only if certain. Crowded blocks also drop weak prefix-only pairs and require stronger Jaro--Winkler ($\geq 0.97$).

\subsubsection{Potential meet merges}
\label{app:meet-merges}

Suggestions require the same sport, similar meet name (exact normalized match or Jaro--Winkler $\geq 0.90$), and start/end dates within 31 days of each other. Rejecting a pair hides it from future suggestions. Meet merges combine result rows under one meet identity.

\subsubsection{Potential meet series}
\label{app:meet-series}

Meet series link the same invite across years so Pros can toggle editions and view series records. This does \emph{not} combine results (unlike meet merges). Suggestions need the same sport, matching names (exact full name, or Jaro--Winkler $\geq 0.90$~\cite{christen2012data} on the distinctive core after stripping shared suffixes such as ``Invitational'' / ``Regional''), and different calendar years. Fuzzy groups require every pair to meet the threshold (no transitive chaining) and at most one edition per year. Regionals for different regions (e.g., Northeast vs.\ Great Lakes) never match. Nationals meets are excluded here---they already get year toggles and series records automatically from the nationals flag (same sport). Meets on the potential meet-merge list (or already queued to merge) are also excluded so near-duplicates are resolved as merges first. Dismiss rejects every pair among the listed editions and any other editions already linked in the same series, so an outsider is suppressed against the whole series (not just the one year that showed up).

\subsection{Historical migration (2003--July 2023)}
\label{app:historical-migration}

Roughly half of export rows predate the comprehensive metadata era. Historical meets were ingested from earlier NRCD/NIRCA digital archives and volunteer-transcribed public results, then passed through the same admin-approval gate as contemporary submissions. Course weather and measured-distance fields were rarely available at ingestion time, so those joins remain sparse by design (Table~\ref{tab:coverage} in the main text; Table~\ref{tab:app-missingness} here). We do not back-impute historical weather from modern APIs unless a meet city, date, and start time are trusted; silent backfill would invent precision the archive never had. Users should stratify analyses by era or restrict environmental models to comprehensive-era rows.

\subsection{Community submission error rates and selection}
\label{app:submission-bias}

Every released row is admin-approved against public postings, which bounds gross fabrication but does not eliminate transcription typos (swapped digits, wrong event assignment). We do not claim a measured residual error rate: no second independent census of all NIRCA finishes exists against which to compute precision/recall. Selection into NRCD is non-random: active clubs and volunteers contribute more meets; inactive programs and small invitationals are under-represented. Admin outreach partially offsets this (nationwide meet addition), but coverage remains NIRCA-centric rather than a census of all U.S.\ collegiate club running. Historical rows inherit whatever meets were archived, amplifying early-adopter bias. Downstream papers should report era filters and avoid treating missing clubs as structural zeros.

\section{Longitudinal Depth by Sport}
\label{app:longitudinal}

Longitudinal modeling requires repeated athlete-seasons. Table~\ref{tab:app-longitudinal} summarizes school-year spans (August~1 boundary) and within-season meet/result counts.

\begin{table}[t]
  \centering
  \footnotesize
  \setlength{\tabcolsep}{2pt}
  \resizebox{\columnwidth}{!}{%
  \begin{tabular}{lrrrrrr}
    \toprule
    Sport & Athletes & $\geq$2 yr & P50 meets & P90 meets & P50 res. & P90 res. \\
    \midrule
    Cross Country   & 23,565 & 36.7\% & 2 & 4 & 2 & 4 \\
    Outdoor Track   &  9,976 & 30.8\% & 1 & 3 & 2 & 5 \\
    Indoor Track    &  9,657 & 33.4\% & 1 & 2 & 1 & 4 \\
    Road Race       &  1,630 & 20.4\% & 1 & 1 & 1 & 1 \\
    \midrule
    \textbf{All sports} & 31,351 & 37.0\% & n/a & n/a & n/a & n/a \\
    \bottomrule
  \end{tabular}%
  }
  \caption{Per sport: athletes with results, \% with $\geq$2 school years, and within-season meets/results per athlete-season (medians and P90). All-sports row from full export.}
  \label{tab:app-longitudinal}
\end{table}

\textbf{Interpretation.} XC offers the densest repeated-meet structure (median 2 meets/season, P90 4) and the highest multi-year retention among sports (36.7\% of XC athletes with $\geq$2 school years; 17.3\% with $\geq$3). Track athletes accumulate more \emph{results} per season (multiple events per meet) but fewer \emph{meets}. Road racing is sparse (median one result per athlete-season).

\section{Public Export Structure and Zenodo Release}
\label{app:dataset-structure}

The analysis in this paper uses the \textbf{approved public CSV export} (v2.1.0, August 2026). Zenodo (\url{https://zenodo.org/records/22077479}) is permanently linked to the GitHub source repository \url{https://github.com/National-Running-Club-Database/national_running_club_database_public_dataset}; each Zenodo release is a snapshot of that repo. The live submission site (\url{https://nationalrunningclubdatabase.com}) accepts new meets continuously. Tables~\ref{tab:app-schema} to~\ref{tab:app-columns-2} summarize tables, joins, and \emph{key fields} as documented in \path{DATASHEET.md} and \path{README.md} on GitHub~\cite{gebru2021datasheets}; full CSV schemas are in \path{README.md}, not repeated here.

\begin{table*}[t]
  \centering
  \small
  \setlength{\tabcolsep}{4pt}
  \begin{tabular}{llp{0.52\textwidth}}
    \toprule
    Table & Primary key & Role \\
    \midrule
    \texttt{result.csv} & \texttt{result\_id} & Finisher time, athlete, meet, event \\
    \texttt{meet.csv} & \texttt{meet\_id} & Date, sport, venue altitude, nationals flag \\
    \texttt{athlete.csv} & \texttt{athlete\_id} & Gender; \emph{no names in public export} \\
    \texttt{team.csv} & \texttt{team\_id} & Institution / club label \\
    \texttt{athlete\_team\_association.csv} & (\texttt{athlete\_id}, \texttt{team\_id}) & Roster membership by season \\
    \texttt{running\_event.csv} & \texttt{running\_event\_id} & Event name and distance \\
    \texttt{sport.csv} & \texttt{sport\_id} & XC, indoor/outdoor track, road \\
    \texttt{course\_details.csv} & (\texttt{meet\_id}, \texttt{running\_event\_id}, \texttt{gender}) & Weather, grade, measured length \\
    \texttt{joined.csv} & \texttt{result\_id} & Denormalized merge of result, meet, team, athlete, course \\
    \bottomrule
  \end{tabular}%
  \caption{NRCD public export tables (v2.1.0: 143{,}868 approved results, 31{,}351 athletes with $\geq 1$ result, 1{,}423 meets, 190 teams). Only \texttt{approved=True} meets and finishers are released.}
  \label{tab:app-schema}
\end{table*}

\begin{table*}[t]
  \centering
  \small
  \setlength{\tabcolsep}{4pt}
  \begin{tabular}{lp{0.78\textwidth}}
    \toprule
    File & Key fields \\
    \midrule
    \texttt{athlete.csv} &
    \texttt{athlete\_id}, \texttt{gender} \\
    \texttt{athlete\_team\_association.csv} &
    \texttt{athlete\_id}, \texttt{team\_id} \\
    \texttt{meet.csv} &
    \texttt{meet\_name}, \texttt{start\_date}, \texttt{meet\_city}, \texttt{meet\_state}, \texttt{altitude}, \texttt{sport\_id}, \texttt{nirca\_only}, \texttt{regionals}, \texttt{nationals} \\
    \texttt{result.csv} &
    \texttt{athlete\_id}, \texttt{meet\_id}, \texttt{running\_event\_id}, \texttt{result\_time}, \texttt{grade}, \texttt{wind}; relay \texttt{athlete\_id\_2} to \texttt{4} and splits \\
    \texttt{running\_event.csv} &
    \texttt{event\_name} \\
    \texttt{sport.csv} &
    \texttt{sport\_name} \\
    \texttt{team.csv} &
    \texttt{team\_name}, \texttt{region}, \texttt{city}, \texttt{state} \\
    \bottomrule
  \end{tabular}%
  \caption{Public-export key fields (relational tables).}
  \label{tab:app-columns}
\end{table*}

\begin{table*}[t]
  \centering
  \small
  \setlength{\tabcolsep}{4pt}
  \begin{tabular}{lp{0.78\textwidth}}
    \toprule
    File & Key fields \\
    \midrule
    \texttt{course\_details.csv} &
    \texttt{elevation\_gain}, \texttt{elevation\_loss}, \texttt{estimated\_course\_distance}, \texttt{date\_of\_event}, \texttt{temperature}, \texttt{dew\_point}, \texttt{humidity}, \texttt{barometric\_pressure}, \texttt{wind\_speed}, \texttt{weather\_conditions}, \texttt{altitude} \\
    \texttt{joined.csv} &
    Denormalized key fields from result, meet, team, athlete, and course tables; \texttt{wind}, \texttt{result\_time}, \texttt{grade}, relay IDs/splits \\
    \bottomrule
  \end{tabular}%
  \caption{Public-export key fields}
  \label{tab:app-columns-2}
\end{table*}

\textbf{Release filtering:} Direct identifiers (legal names, profile URLs, contact fields) are removed before Zenodo upload. Athletes receive stable pseudonymous \texttt{athlete\_id} values; teams retain institution names because they are already public on meet results pages. DNS/DNF rows are absent; the export is finisher-only.

\textbf{Re-identification risk and out-of-corpus linkage:} Meet times, team, and approximate date are often already \emph{public} on host-university results sites, Milesplit, TFRRS, and NIRCA meet pages linked from the project website. NRCD therefore does not introduce a wholly new class of sensitive facts for most rows; it aggregates and standardizes information that motivated runners and coaches can already look up. Residual risk remains: a motivated party could join \texttt{athlete\_id} across meets within NRCD, or link a small number of distinctive performances back to public web results (\emph{out-of-corpus re-identification}). We treat this as a misuse scenario rather than a product goal: researchers should not attempt to recover identities, publish deanonymized subsets, or merge NRCD with social-media graphs. IRB review classified the public export as non-human-subjects research with PII removed (University of Notre Dame protocol~25-05-9348); downstream work should cite the Zenodo DOI and respect the datasheet use constraints (\path{DATASHEET.md} in the GitHub dataset repository).

\textbf{Recommended acquisition path:} Download the Zenodo archive (\url{https://zenodo.org/records/22077479}) or clone / create a submodule of the GitHub dataset repository (\url{https://github.com/National-Running-Club-Database/national_running_club_database_public_dataset}). Read \path{DATASHEET.md} and \path{README.md} there for column semantics, era splits, and intended uses.

\section{Sustainability, Versioning, and Intended Users}
\label{app:sustainability}

\textbf{Version control.} Public dataset releases are Git-tagged in the dataset repository and mirrored as Zenodo versioned deposits under a concept DOI so citations resolve to a specific snapshot. Analysis code for this paper lives in a separate GitHub repository with tagged releases matching the export used here (v2.1.0). The \texttt{nrcd} package is versioned on PyPI/GitHub independently so formula bug fixes can ship without re-exporting CSVs.

\textbf{Update cadence:} The live site accepts meets continuously; public CSV exports are planned \textbf{yearly} (or on major schema changes), each with an updated datasheet. Meet volume growth (78$\to$145$\to$335$\to$274 approved meets in 2022--2025) is the operational sustainability signal.

\textbf{Succession:} Day-to-day curation is organized around the National Running Club Database LLC rather than a single student author. Admin approval remains a human gate. Any domain-expert admin with repository and site credentials can continue approvals for years to come. Open issues/PRs remain the contribution channel for formulas.

\textbf{Expected users:} (i)~Sports-informatics and data-mining researchers needing longitudinal, gender-balanced officiated results; (ii)~environmental and fairness analysts studying weather/grade confounders; (iii)~coaches and student clubs standardizing private times via \texttt{nrcd}; (iv)~equity researchers examining women's club participation (39.4\% of athletes). The resource is deliberately not a recruiting ranking product.

\section{Empowering Users with the \texttt{nrcd} Python Package}
\label{app:python-package}

Standardization formulas in the main text are implemented in the standalone \textbf{\texttt{nrcd}} package (PyPI: \url{https://pypi.org/project/nrcd/}; source: \url{https://github.com/National-Running-Club-Database/nrcd}), decoupled from the Zenodo CSVs. Athletes, coaches, and researchers can apply the same adjustments to private or newly collected results. API details are in \texttt{PARAMETERS\_DOC} and the package README.

\textbf{Use case 1: fair cross-meet XC comparison.} A woman runs 24:10 on a listed 6K course that actually measures 6.2\,km, with ${\sim}$120\,m of climb at 78\,°F; \texttt{standardize\_xc} converts distance, grade, heat, and altitude to a comparable 6\,km finish time. Distances accept \texttt{km}, \texttt{mi}, or meters; weather defaults to °F or \texttt{temp\_unit="C"}; grade accepts vertical m, feet, or percent:
\begin{figure*}[t]
\centering
\small
\begin{minipage}{0.92\textwidth}
\begin{verbatim}
from nrcd.standardize import format_time, standardize_xc

std = standardize_xc(
    "24:10",
    gender="F",
    reported_distance="6k",        # alt: 6.2, "6.2", reported_distance_m=6000
    distance_unit="km",              # alt: "mi"; or pass *_distance_m in meters
    actual_distance=6.2,             # alt: actual_distance_m=6200
    temperature=78, dew_point=70,  # °F default; alt: temp_unit="C"
    elevation_gain=120,
    elevation_loss=120,
    grade_input="m",                 # alt: "feet", or omit for % grade
    target_distance=6000,            # alt: "6k", "6000m", target_distance_m=6000
)
print(format_time(std))  # e.g. 21:45.67 on standardized 6 km
\end{verbatim}
\end{minipage}
\end{figure*}

\textbf{Use case 2: outdoor wind vs.\ indoor banked track.} \\\texttt{standardize\_outdoor\_track} applies \texttt{wind\_mps} on outdoor sprints only; \texttt{standardize\_indoor\_track} uses \texttt{lap\_length\_m} and \texttt{banked} indoors with no wind term.

\textbf{Use case 3: reproducible research on Zenodo.} Load approved CSVs with the package dataset helpers and regenerate paper statistics from the same formulas and \texttt{StandardizeConfig} coefficients used in Appendix~\ref{app:riegel-sens}.

\textbf{Use case 4: metadata backfill for new meets.} \\ \texttt{enrich\_race\_context} can populate temperature, dew point, and U.S.\ venue altitude from city, state, and race datetime; standardization itself requires no API keys.

The package is MIT-licensed open source on GitHub: open an issue when something is missing or incorrect, or fork and submit pull requests with improvements. Cite the resource paper when publishing work that relies on its formulas.

\section{Factors Not Covered by NRCD Standardization}
\label{app:uncovered}

Researchers designing studies with NRCD should treat the factors below as \emph{potential covariates, stratification variables, or sensitivity analyses}. They are not defects in the export, but are latent modifiers that standardized times do not encode. NRCD adjusts officiated meet times for distance, course length, grade, heat, altitude, and (track) venue and wind where metadata exist; the subsections below cite literature on what remains unmodeled so downstream work can bound or explicitly model it~\cite{riegel1981athletic,maurer2018race}.

\subsection{Training load, aerobic fitness, and recovery markers}
\label{app:uncovered-training}

\textbf{Training volume and periodization:} Weekly mileage, intensity distribution, taper length, and recency of hard sessions all influence race readiness on a given weekend~\cite{seiler2010best}. Longer horizons matter too: summer base miles, off-season work, and pre-season preparation predict in-season performance in collegiate distance runners, not just the prior week~\cite{lundstrom2025pre}. Meta-analytic evidence supports structured tapers: about two weeks with 41 to 60\% volume reduction while maintaining intensity yields measurable performance gains~\cite{bosquet2007effects}.

\textbf{Aerobic capacity:} Meet results do not encode lab- or field-derived fitness such as $\dot{V}\mathrm{O}_{2\max}$, running economy, or lactate and ventilatory thresholds, even though these traits explain much of the spread in endurance performance~\cite{besson2022sex}. Standardized times, therefore, conflate a one-day outcome with an unobserved aerobic profile built over months of training.

\textbf{Heart-rate variability:} HRV from wearables or morning resting measures is used to flag autonomic recovery and training readiness across multi-week blocks~\cite{rothschild2024predicting}. NRCD has no HRV, sleep-stage, or training-log fields, so day-to-day freshness is latent even when meet dates are known.

\textbf{Body composition:} Seasonal changes in body mass and lean mass shift running economy even when aerobic fitness is stable~\cite{hoogkamer2016altered}. NRCD does not record weight, body-fat percentage, or composition scans.

NRCD records \emph{when} athletes raced, not \emph{how} they trained beforehand, their underlying aerobic capacity, or their recovery state that week; two athletes with identical standardized times may differ sharply in summer training load, $\dot{V}\mathrm{O}_{2\max}$, HRV, glycogen stores, and neuromuscular readiness.

\subsection{Within-season race frequency and cumulative fatigue}
\label{app:uncovered-race-frequency}

Meet density over the prior two to three weeks is distinct from chronic training load: repeated competitions compress recovery even when weekly mileage is unchanged~\cite{howle2020injury,page2023effects}. NRCD lists every finisher time but not how many races an athlete started in the surrounding fortnight, so short-term staleness and fixture congestion are latent when interpreting a single slow result or a within-season improvement slope (Section~\ref{app:improvement-direction}).

\subsection{Injury and illness}
\label{app:uncovered-injury}

Acute and overuse injuries alter biomechanics, limit training continuity, and often appear in meet data as DNS/DNF rather than slow finish times~\cite{nielsen2014prospective,nielsen2014excessive}. Since NRCD only records finish times, this is not taken into account. Prospective cohort studies link rapid weekly distance progression (for example, $>$30\% week over week) to elevated hazard of patellofemoral pain, iliotibial band syndrome, and related distance-running injuries~\cite{nielsen2014excessive}. Upper-respiratory illness, concussion return-to-play protocols, and subclinical illness similarly decouple observed performance from underlying capacity. Without injury logs or training diaries, latent health state remains unobserved.

\subsection{Sleep}
\label{app:uncovered-sleep}

Sleep loss is common before competition and can impair sport-specific performance, cognitive function, and autonomic balance even when maximal strength tasks are preserved~\cite{fullagar2015sleep}. Partial sleep restriction elevates perceived exertion and slows decision-relevant processing, both relevant to mass-start XC and tactical track races. NRCD has no sleep-duration or sleep-quality fields; late-night travel, exam weeks, and early bus departures for away meets are invisible confounders.

\subsection{Nutrition and energy availability}
\label{app:uncovered-nutrition}

Carbohydrate availability, pre-race fueling, and day-of-meet energy intake influence glycogen stores, perceived exertion, and late-race pace regulation. Joint position statements recommend individualized plans for macronutrient timing, recovery nutrition, and safe supplementation~\cite{thomas2016nutrition}. Low energy availability, whether from intentional restriction, poor meal access on travel weekends, or academic schedule disruption, can impair performance independently of fitness and overlaps with RED-S pathways in women (Section~\ref{app:uncovered-reds}) and men (Section~\ref{app:uncovered-men})~\cite{mountjoy2018ioc}. NRCD records no dietary logs, body mass on race day, or supplement use; two athletes with identical standardized times may differ in prerace carbohydrate loading or micronutrient status (e.g.\ iron, vitamin D).

\subsection{Caffeine and legal ergogenic aids}
\label{app:uncovered-caffeine}

Caffeine is widely used in collegiate racing and can improve endurance performance at moderate doses, independent of weather and grade corrections~\cite{guest2021international,thomas2016nutrition}. Beetroot nitrate, sodium bicarbonate, and other legal aids show smaller or event-specific effects in position-stand summaries~\cite{thomas2016nutrition}. NRCD stores no supplement timing or dose fields, so identical standardized times may reflect different prerace ergogenic protocols.

\subsection{Hydration, dehydration, and exercise-associated cramping}
\label{app:uncovered-hydration}

Fluid deficits can raise cardiovascular strain and degrade endurance performance in warm conditions, especially when athletes under-drink relative to sweat loss~\cite{thomas2016nutrition}. Exercise-associated muscle cramping (EAMC) during hard mass-start XC is often attributed to dehydration or sodium loss, but prospective endurance studies find cramping more closely associated with higher running intensity, prior cramp history, and neuromuscular fatigue than with abnormal serum electrolytes or greater body-mass loss at the time of cramp~\cite{schwellnus2009cause,schwellnus2011increased,nelson2016narrative}. NRCD does not record fluid intake or cramp events; a single slow finish time cannot distinguish cramp-shortened effort from fitness limits. Heat standardization adjusts expected thermal stress but not hydration behavior.

\subsection{Travel and jet lag}
\label{app:uncovered-travel}

Long-haul travel induces travel fatigue (discomfort, mild hypoxia, disrupted routines) and, when crossing multiple time zones, jet lag (circadian misalignment, sleep disruption, gastrointestinal symptoms)~\cite{janse2021managing}. Controlled studies report impaired sprint and intermittent endurance performance for several days after eastward transmeridian flights, with larger effects than westward travel of equal duration~\cite{fowler2017greater}. Even same-day car trips to regional meets can shorten sleep and disrupt prerace fueling~\cite{janse2021managing,thomas2016nutrition}. Travel therefore remains a latent performance modifier for away meets throughout the season. XC validation and within-season improvement summaries in the main text \textbf{exclude NIRCA nationals}, which \emph{partially} limits how much travel confounds those checks; it does not remove travel effects on regular-season results. NRCD encodes meet and team location but not hours in transit or arrival time relative to race start.

\subsection{Academic load and student-athlete time demands}
\label{app:uncovered-academics}

NCAA student-athletes report median in-season commitments near 66.5 hours per week combined across athletics and academics~\cite{ncaa2025goals}, with exam periods, lab deadlines, and missed class time adding stress independent of training load. Systematic reviews document elevated depression and anxiety prevalence relative to non-athlete students, often under-reported because of stigma and performance culture~\cite{beisecker2024depression}. Cognitive fatigue and all-nighter study sessions can blunt race-day effort regulation even when physical training is unchanged. NRCD links athletes to institutions but not to course schedules, GPA pressure, or academic calendar events. Additionally, several NIRCA  clubs are student run, meaning that certain runners also need to deal with race-day logistics.

\subsection{Circadian rhythm and time of day}
\label{app:uncovered-circadian}

Morning versus afternoon start times shift core temperature, hormone profiles, and perceived effort even when weather metadata are held constant~\cite{kang2023time}. \texttt{time\_of\_event} is stored for comprehensive-era rows, but standardization does not adjust for circadian phase; a cool 8:00\,am race and a warm 4:00\,pm race on the same course are not equated for body-clock effects. Travel-related circadian disruption is covered separately (Section~\ref{app:uncovered-travel}).

\subsection{Precipitation, footing, and solar load}
\label{app:uncovered-precipitation}

Rain, snow, mud, and saturated grass change grip and energy cost in ways that temperature and dew point alone do not capture~\cite{lejeune1998mechanics,pinnington2001energy,wilson2024assessing}. Comprehensive-era exports may include precipitation fields from OpenWeather, but the standardization pipeline does not apply a footing factor; wet-course effects remain in the residual (Section~\ref{app:uncovered-footwear}). Direct solar radiation and UV exposure on open courses add radiant heat load beyond the heat-index surrogate used for air temperature and humidity~\cite{brotherhood2008heat,zhao2015effect}. Cloud cover and UV index are sometimes present in the course details but are not part of the published adjustment formulas.

\subsection{Altitude acclimatization}
\label{app:uncovered-acclimatization}

Venue altitude enters through P\'eronnet-style corrections (main text), but days at elevation before the race are not modeled~\cite{peronnet1991theoretical,chapman2010altitude}. An athlete who lives or trains at altitude for weeks differs from one who drives to a mountain meet the morning of the race, even at the same starting line and barometric pressure. NRCD encodes meet elevation, not acclimatization history.

\subsection{Footwear and surface conditions}
\label{app:uncovered-footwear}

Shoe mass, cushioning, plate stiffness, and spike length alter economy and traction; roughly 1\% higher metabolic cost per 100\,g added mass slows time trials proportionally~\cite{hoogkamer2016altered}. Compliant surfaces (mud, saturated grass, snow, ice) dissipate elastic energy and raise stabilizer demand~\cite{lejeune1998mechanics,pinnington2001energy,voloshina2015biomechanics,macdermid2021comparative,hollis2021running}. Precipitation-specific effects are discussed in Section~\ref{app:uncovered-precipitation}. Meet results record neither footwear model nor spike choice nor post-race soil moisture.

\subsection{Gender-specific factors in women}
\label{app:uncovered-women}

NRCD standardizes times with gender-specific Riegel exponents and separate XC target distances (6\,km women, 8\,km men), but does \emph{not} encode physiology that disproportionately affects women. For equity research using NRCD, within-gender modeling is essential; pooling or comparing raw seconds across genders without distance conversion remains inappropriate. The subsections below summarize latent factors documented primarily in women that can shift race-day performance or race availability independently of weather, grade, or altitude.

\subsubsection{Menstrual-cycle phase and endogenous hormones}
\label{app:uncovered-menstrual-phase}

Fluctuations in oestradiol and progesterone across the menstrual cycle can alter substrate use, thermoregulation, and perceived exertion. A 2020 meta-analysis of eumenorrheic women found a \emph{trivial} average reduction in endurance and strength performance in the early follicular phase relative to other phases (median pooled effect $\approx -0.06$), with low study quality and large between-study heterogeneity~\cite{mcnulty2020effects}. The largest phase contrast (early vs.\ late follicular) remained small (effect $\approx -0.14$). Population guidelines by cycle phase are therefore not supported; individual tracking is recommended when performance matters. NRCD stores no cycle phase, hormonal contraceptive status, or symptom diary fields, so cycle-aligned meet dates are an unobserved confounder for within-woman season trajectories.

\subsubsection{Menstrual symptoms and race availability}
\label{app:uncovered-menstrual-symptoms}

Cycle-related symptoms can affect performance even when laboratory tests of ``phase'' show only trivial mean shifts. In a survey of 6,812 exercising women (Strava cohort), 84 to 91\% reported cramps, fatigue, mood changes, or breast pain across the cycle; higher symptom burden predicted missing or altering training and missing competitions (odds ratios $\approx 1.07 to 1.09$ per unit on a 54-point symptom index)~\cite{bruinvels2021prevalence}. For NRCD, this matters in two ways: (i) a slow time may reflect symptom-limited effort rather than fitness, and (ii) women who skip meets during symptomatic weeks are \emph{censored from the dataset} (finisher-only export), biasing observed improvement rates and participation counts. DNS/DNF rows are absent, so symptom-driven non-starts are invisible.

\subsubsection{Iron deficiency and anemia}
\label{app:uncovered-iron-women}

Iron deficiency is more prevalent in female than male endurance athletes (often cited 15 to 35\% vs.\ 3 to 11\% in training populations) because of menstrual losses, lower dietary intake, foot-strike hemolysis, and exercise-induced inflammation~\cite{sinclair2005prevalence,burden2015iron}. Even without anemia, low ferritin can reduce $\dot{V}\mathrm{O}_{2\max}$ and increase fatigue~\cite{burden2015iron}. NRCD has no ferritin, hemoglobin, or supplementation fields; two women with identical standardized XC times may differ in oxygen-carrying capacity.

\subsubsection{Low energy availability and menstrual dysfunction ({RED-S})}
\label{app:uncovered-reds}

Relative energy deficiency in sport (RED-S) arises when energy intake does not cover exercise and physiological needs. It impairs metabolic rate, bone health, immunity, cardiovascular function, and menstrual function in women~\cite{mountjoy2018ioc}. Functional hypothalamic amenorrhoea and oligomenorrhoea are common warning signs in female distance runners and can precede stress fractures and prolonged performance plateaus. RED-S is not captured by meet times; affected athletes may race poorly, reduce start frequency, or leave the sport entirely. Coaches and clinicians screen with energy-availability and menstrual-history tools that NRCD does not contain.

\subsubsection{Oral contraceptives and hormonal contraception}
\label{app:uncovered-ocp}

Many collegiate women use combined oral contraceptives (OCPs), which suppress ovulation and create ``pseudo-phases'' of synthetic estrogen and progestin distinct from the natural cycle~\cite{elliott2020effects}. A 2020 meta-analysis found OCP users may perform slightly worse on average than naturally menstruating women, but the group effect is trivial and highly variable; performance was stable across OCP consumption vs.\ withdrawal weeks~\cite{elliott2020effects}. NRCD does not record contraceptive type or timing, so researchers cannot stratify by hormonal profile.

\subsubsection{Implications for women's XC in NRCD}
\label{app:uncovered-women-implications}

Women constitute 39.4\% of unique athletes in the NRCD export (12,364 women vs.\ 18,987 men; 36.2\% of result rows) and show larger raw-to-standardized variance compression than men (51.1\% vs.\ 35.4\% median cross-meet SD reduction), driven mainly by distance normalization rather than larger weather swings. Gender-stratified validation in the main resource reflects this. Analyses that treat standardized times as complete fitness proxies for women should explicitly acknowledge menstrual physiology, iron status, energy availability, and symptom-driven meet absence as unmeasured moderators~\cite{mcnulty2020effects,bruinvels2021prevalence,mountjoy2018ioc}.

\subsection{Latent factors in men's physiology}
\label{app:uncovered-men}

Men constitute 60.6\% of unique athletes in the export and 63.8\% of result rows, yet meet times do not encode male-specific physiology either. Endogenous testosterone and training-response differences between men and women are summarized in recent consensus work~\cite{hunter2023biological,besson2022sex}. Iron deficiency without anemia is common in recreationally active men, not only women, and can limit aerobic capacity~\cite{sinclair2005prevalence,burden2015iron}. Low energy availability and RED-S also occur in male distance runners (bone stress, hormonal suppression, performance decline) but are under-screened relative to women's sport~\cite{mountjoy2018ioc,blauwet2017low}. Gender-stratified modeling remains essential; these gaps mean standardized men's times still confound latent health and hormonal state with environment-adjusted performance.

\subsection{Psychology, pacing, and race tactics}
\label{app:uncovered-pacing}

Effort is regulated anticipatorily from expected duration, prior experience, and afferent feedback~\cite{tucker2009physiological}; championship stakes, team-scoring position, and rivalry meets further shift risk tolerance and early surges. Running in a pack reduces effective wind resistance and aerobic demand compared with leading or racing alone~\cite{pugh1971influence,hettinga2019science}, distinct from sit-and-kick or surge tactics. Mass-start XC is tactically heterogeneous: World Championships analyses show positive (fade) pacing for most finishers, with medalists separating only late~\cite{hanley2014senior,hanley2018pacing}. Hill surges, lane congestion, and mid-race position change instantaneous speed in ways we cannot capture~\cite{hettinga2019science}. NRCD stores one aggregate time per result, not splits, pack position, or within-race tactical profile. Field depth, course familiarity, and home-context effects are selection issues (Section~\ref{app:uncovered-selection}).

\subsection{Selection and roster composition}
\label{app:uncovered-selection}

Who appears in NRCD depends on club vs.\ varsity thresholds at a given university, athletes choosing to DNS, and self-selection~\cite{anderson2023under,james2023underrepresentation}. A fast time from a thin roster meet is not exchangeable with a depth-loaded invitational. Competitive field quality shifts early pace and risk taking even when an athlete's fitness is unchanged~\cite{hanley2018pacing,hettinga2019science}. Athletes who train on or know a course, or compete near home with familiar travel and support, may gain an edge not captured by weather and grade metadata~\cite{courneya1991effects,wilson2024assessing}. Meet-level \texttt{regionals} and \texttt{nationals} flags (Section~\ref{app:dataset-structure}) mark post-season championship context but do not encode race-day hype: crowd energy, broadcast pressure, and ``big-meet'' arousal can lift or disrupt performance beyond weather and grade adjustments, and often align with the nationals-heavy career PRs in Section~\ref{app:xc-percentiles}. Club runners sometimes sacrifice their own effort to pace a teammate toward a PR or fill a supporting team-scoring role; that choice is invisible in one finisher time and can bias within-season improvement rates (Section~\ref{app:improvement-direction}), separate from the within-race tactics in Section~\ref{app:uncovered-pacing}. Spectators (family, friends, and home-club supporters) can further raise arousal and perceived support, interacting with academic stress (Section~\ref{app:uncovered-academics}) and away-meet travel (Section~\ref{app:uncovered-travel}). Standardization cannot correct for who chose to start, if they got excited for nationals, or whether they raced for a personal outcome vs.\ a teammate's PR.

\subsection{Course geometry beyond aggregate grade}
\label{app:uncovered-course}

Recorded gain/loss and measured length summarize certain course features, but not turn frequency, terrain, or choke points. Uneven micro-terrain raises energy cost by several percent even on modest height variation~\cite{voloshina2015biomechanics}; footing effects (Section~\ref{app:uncovered-footwear}) can add further separation between courses with identical elevation metadata.

\subsection{Officiating, timing, and wind on non-sprint track events}
\label{app:uncovered-officiating}

Photo-finish latency, course-cutting penalties, and starter variability introduce measurement error unrelated to fitness. Wind readings exist for sprints and jumps where recorded, but middle-distance track races often lack per-lane wind metadata; non-sprint wind corrections are not applied in NRCD~\cite{linthorne1994effect,quinn2003effects}.

\textbf{Researchers using NRCD for causal or predictive modeling should treat standardized times as environment-adjusted outcomes and explicitly model or bound the latent factors above.}